\documentclass[twoside,11pt]{article}
\pdfoutput=1

\usepackage{geometry}
\usepackage{natbib}
\usepackage{graphicx}
\usepackage{rotating}

\usepackage{amsmath}
\usepackage{amssymb}

\usepackage{theorem}
\newcommand{\BlackBox}{\rule{1.5ex}{1.5ex}}  
\newenvironment{proof}{\par\noindent{\bf Proof\ }}{\hfill\BlackBox\\[2mm]}

\newtheorem{theorem}{Theorem}

\newtheorem{lemma}[theorem]{Lemma}

\newcommand{\eq}[1]{(\ref{#1})}
\newcommand{\mymatrix}[2]{\left[\begin{array}{#1} #2 \end{array}\right]}

\newcommand{\inner}[2]{\left\langle #1,#2 \right\rangle}
\newcommand{\rbr}[1]{\left(#1\right)}
\newcommand{\sbr}[1]{\left[#1\right]}
\newcommand{\cbr}[1]{\left\{#1\right\}}
\newcommand{\nbr}[1]{\left\|#1\right\|}
\newcommand{\abr}[1]{\left|#1\right|}

\newcommand{\RR}{\mathbb{R}}

\newcommand{\Xcal}{\mathcal{X}}

\newcommand{\Ycal}{\mathcal{Y}}

\newcommand{\Ncal}{\mathcal{N}}

\newcommand{\Pb}{\mathbf{P}}

\DeclareMathOperator*{\argmax}{\mathrm{argmax}}
\DeclareMathOperator*{\argmin}{\mathrm{argmin}}
\DeclareMathOperator*{\sgn}{\mathrm{sgn}}
\DeclareMathOperator*{\tr}{\mathrm{tr}}
\DeclareMathOperator*{\cov}{\mathrm{Cov}}
\DeclareMathOperator*{\var}{\mathrm{Var}}
\DeclareMathOperator*{\mini}{\mathrm{minimize}}
\DeclareMathOperator*{\maxi}{\mathrm{maximize}}

\newcommand{\intset}[1]{\cbr{1..n}}
\newcommand{\flexset}[2]{\cbr{{#1}_1, \ldots, {#1}_{#2}}}

\usepackage[colorlinks,pdfpagelabels,plainpages=false]{hyperref}
\usepackage{algorithm}
\usepackage{algorithmic}
\usepackage{rotating}

\usepackage{xcolor}
\definecolor{dark-red}{rgb}{0.4,0.15,0.15}
\definecolor{dark-blue}{rgb}{0.15,0.15,0.4}
\definecolor{medium-blue}{rgb}{0,0,0.5}
\hypersetup{
   colorlinks, linkcolor={dark-blue},
   citecolor={dark-blue}, urlcolor={medium-blue}
}

\graphicspath{{../../Figures/}}

\usepackage{array}
\newcolumntype{L}[1]{>{\raggedright\let\newline\\\arraybackslash\hspace{0pt}}m{#1}}
\newcolumntype{C}[1]{>{\centering\let\newline\\\arraybackslash\hspace{0pt}}m{#1}}
\newcolumntype{R}[1]{>{\raggedleft\let\newline\\\arraybackslash\hspace{0pt}}m{#1}}

\newcommand{\gp}{\mathrm{GP}}

\newcommand{\bm}{}
\newcommand{\FIXME}[1]{\textcolor{red}{[#1]}}
\usepackage{comment}
\usepackage{subcaption}

\title{{A} la Carte --- Learning Fast Kernels}
\author{
  Zichao Yang \\
  Carnegie Mellon University, 5000 Forbes Ave, Pittsburgh 15213
  PA, USA
  \and
  Alexander J. Smola \\
  Carnegie Mellon University, 5000 Forbes Ave, Pittsburgh 15213
  PA, USA \\
  Google~Strategic Technologies, 1600 Amphitheatre Pky, Mountain View 94043 CA, USA
  \and
  Le Song \\
  Georgia Institute of Technology, 1340 Klaus Drive, Atlanta 30332 GA, USA
  \and
  Andrew Gordon Wilson \\
  Carnegie Mellon University, 5000 Forbes Ave, Pittsburgh 15213
  PA, USA
}

\begin{document}
\date{\today}

\maketitle

\begin{abstract}
Kernel methods have great promise for learning rich statistical
representations of large modern datasets.  However, compared to
neural networks, kernel methods have been perceived as lacking
in scalability and flexibility.
We introduce a family of fast, flexible, lightly parametrized
and general purpose kernel learning methods, derived
from Fastfood basis function expansions. We provide mechanisms
to learn the properties of groups of spectral frequencies in these
expansions, which require only $\mathcal{O}(m \log d)$ time and
$\mathcal{O}(m)$ memory, for $m$ basis functions and $d$ input
dimensions.  We show that the proposed methods can learn a wide class
of kernels, outperforming the alternatives in accuracy,
speed, and memory consumption.

\end{abstract}

\section{Introduction}

The generalisation properties of a kernel method are entirely
controlled by a kernel function, which represents an inner
product of arbitrarily many basis functions.  Kernel methods
typically face a tradeoff between speed and flexibility.
Methods which learn a kernel lead to slow and expensive to
compute function classes,
whereas many fast function classes are not adaptive.  This problem
is compounded by the fact that expressive kernel learning
methods are most needed on large modern datasets, which
provide unprecedented opportunities to automatically learn rich statistical
representations.

For example, the recent spectral kernels proposed by
\citet{WilAda13} are flexible, but require an arbitrarily
large number of basis functions, combined with many free
hyperparameters, which can lead to major computational
restrictions.
Conversely, the recent Random Kitchen Sinks of \citet{RahRec09} and
Fastfood \citep{LeSarSmo13} methods offer efficient finite basis
function expansions, but only for \emph{known} kernels, a priori
hand chosen by the user.  These methods do not address the
fundamental issue that it is exceptionally difficult to know a-priori which
kernel might perform well; indeed, an appropriate kernel might
not even be available in closed form.

We introduce a family of kernel learning methods
which are expressive, scalable, and general purpose. In particular,
we introduce flexible kernels, including a novel piecewise
radial kernel, and derive Fastfood basis function expansions for these kernels.
We observe that the frequencies in these expansions can in fact be adjusted,
and provide a mechanism for automatically learning these frequencies via
marginal likelihood optimisation. Individually adjusting these frequencies
provides the flexibility to learn any translation invariant kernel.
However, such a procedure has as many free parameters as basis functions,
which can lead to over-fitting, troublesome local optima, and computational
limitations.  We therefore further introduce algorithms which can control
the scales, spread, and locations of \emph{groups} of frequencies.  These
methods are computationally efficient, and allow for great flexibility,
with a minimal number of free parameters requiring training.
By controlling groups of spectral frequencies, we can use arbitrarily
many basis functions with no risk of over-fitting.  Furthermore, these
methods do not require the input data have any special structure
(e.g., regular sampling intervals).

Overall, we introduce four new
kernel learning methods with distinct properties, and evaluate each of
these methods on a wide range of real datasets.  We show major
advantages in accuracy, speed, and memory consumption.
We begin by describing related work in more detail in
section~\ref{sec:related}.  We then provide additional
background on kernel methods, including basic
properties and Fastfood approximations, in
section~\ref{sec:kernels}.  In section~\ref{sec:alacarte} we introduce a number of
new tools for kernel learning.
Section~\ref{sec:experiments} contains an evaluation of the proposed
techniques on many datasets.   We conclude with a
discussion in section~\ref{sec:blah}.

\section{Related Work}
\label{sec:related}

\citet{RahRec08} introduced \emph{Random Kitchen Sinks} finite Fourier basis
function approximations to \emph{fixed} stationary kernels, using a Monte Carlo sum
obtained by sampling from spectral densities.  For greater flexibility, one can
consider a weighted sum of random kitchen sink expansions of \citet{RahRec09}.
In this case, the expansions are fixed, corresponding to a-priori chosen kernels,
but the weighting can be learned from the data.  

Recently,
\citet{LuMayLiuGaretal14} have shown how weighted sums of
random kitchen sinks can be incorporated into scalable logistic
regression models.
First, they separately learn the parameters of multiple logistic
regression models, each of which uses a separate random
kitchen sinks expansion, enabling parallelization.  They then
jointly learn the weightings of each expansion.  Learning
proceeds through stochastic gradient descent.
\citet{LuMayLiuGaretal14} achieve
promising performance on acoustic modelling
problems, in some instances outperforming deep neural networks.

Alternatively, \citet{Lazaroetal10} considered optimizing
the locations of all spectral frequencies in Random Kitchen Sinks
expansions, as part of a sparse spectrum Gaussian process formalism
(SSGPR).

For further gains in scalability,
\citet{LeSarSmo13} approximate the sampling step in Random Kitchen
Sinks by a combination of matrices which enable fast computation.
The resulting \emph{Fastfood} expansions perform similarly to Random
Kitchen Sinks expansions \citep{LeSarSmo13}, but can be computed
more efficiently.  In particular, the Fastfood expansion requires
$\mathcal{O}(m \log d)$ computations and $\mathcal{O}(m)$ memory,
for $m$ basis functions and $d$ input dimensions.

To allow for highly flexible kernel learning, \citet{WilAda13} proposed
spectral mixture kernels, derived by modelling a spectral density by a
scale-location mixture of Gaussians.  These kernels can be computationally
expensive, as they require arbitrarily many basis functions combined with
many free hyperparameters.  Recently, \citet{WilGilNehCun14} modified
spectral mixture kernels for Kronecker structure, and generalised scalable
Kronecker (Tensor product) based learning and inference procedures
to incomplete grids.  Combining these kernels and inference procedures
in a method called \emph{GPatt}, \citet{WilGilNehCun14} show how to learn rich statistical
representations of large datasets with kernels, naturally enabling
extrapolation on problems involving images, video, and spatiotemporal
statistics.  Indeed the flexibility of spectral mixture kernels makes them
ideally suited to large datasets.  However, GPatt requires that the input
domain of the data has at least partial grid structure in order to see
efficiency gains.

In our paper, we consider weighted mixtures of Fastfood expansions,
where we propose to learn \emph{both} the weighting of the
expansions \emph{and} the properties of the expansions themselves.
We propose several approaches under this framework.  We consider
learning all of the spectral properties of a Fastfood expansion.
  We also
consider learning the properties of groups of spectral frequencies, for
lighter parametrisations and useful inductive biases, while retaining
flexibility.  For this purpose, we show how to incorporate Gaussian
spectral mixtures into our framework, and also introduce novel piecewise
linear radial kernels.  Overall, we show how to perform simultaneously flexible and
scalable kernel learning, with interpretable, lightly parametrised and general purpose
models, requiring no special structure in the data.
We focus on regression for 
clarity, but our models extend to classification and non-Gaussian
likelihoods without additional methodological innovation.

\section{Kernel Methods}
\label{sec:kernels}

\subsection{Basic Properties}
\label{sec:basic}

Denote by $\Xcal$ the domain of covariates and by $\Ycal$ the domain of
labels. Moreover, denote $X := \cbr{x_1, \ldots, x_n}$ and
$Y :=\cbr{y_1, \ldots, y_n}$ data drawn from a joint distribution $p$ over
$\Xcal \times \Ycal$. Finally, let $k: \Xcal \times \Xcal \to \RR$ be a Hilbert
Schmidt kernel \citep{Mercer09}. Loosely speaking we require that $k$ be
symmetric, satisfying that every matrix $K_{ij} := k(x_i, x_j)$ be positive
semidefinite, $K \succeq 0$.

The key idea in kernel methods is that they allow one to represent inner
products in a high-dimensional feature space implicitly using
\begin{align}
  k(x, x') = \inner{\phi(x)}{\phi(x')}.
\end{align}
While the existence of such a mapping $\phi$ is guaranteed by the
theorem of
\citet{Mercer09}, manipulation of $\phi$ is not generally desirable since it
might be infinite dimensional. Instead, one uses the representer theorem
\citep{KimWah70,SchHerSmo01} to show that when solving regularized risk
minimization problems, the optimal solution $f(x) = \inner{w}{\phi(x)}$ can be
found as linear combination of kernel functions:
\begin{align}
  \nonumber
  \inner{w}{\phi(x)} = \inner{\sum_{i=1}^n \alpha_i
    \phi(x_i)}{\phi(x)}
  = \sum_{i=1}^n \alpha_i k(x_i, x).
\end{align}
While this expansion is beneficial for small amounts of data, it creates an
unreasonable burden when the number of datapoints $n$ is large. This problem
can be overcome by computing approximate expansions.

\subsection{Fastfood}
\label{sec:Fastfood}

The key idea in accelerating $\inner{w}{\phi(x)}$ is to find an explicit
feature map such that $k(x,x')$ can be approximated by
$\sum_{j=1}^m \psi_j(x) \psi_j(x')$ in a manner that is both fast and memory
efficient. Following the spectral approach proposed by \citet{RahRec09} one
exploits that for translation invariant kernels $k(x,x') = \kappa(x-x')$ we
have
\begin{align}
  \label{eq:transinv}
  k(x,x') = \int \rho(\omega) \exp\rbr{i \inner{\omega}{x-x'}} d\omega \,.
\end{align}
Here $\rho(\omega) = \rho(-\omega) \geq 0$ to ensure that the imaginary parts
of the integral vanish. Without loss of generality we assume that
$\rho(\omega)$ is normalized, e.g.\ $\nbr{\rho}_1 = 1$. A similar spectral
decomposition
holds for inner product kernels $k(x,x') = \kappa(\inner{x}{x'})$
\citep{LeSarSmo13,Schoenberg42}.

\citet{RahRec09} suggested to sample from the spectral distribution
$\rho(\omega)$ for a Monte Carlo approximation to the integral in \eqref{eq:transinv}.
For example, the Fourier
transform of the popular Gaussian kernel is also Gaussian, and thus samples from a normal distribution for $\rho(\omega)$ can be used to approximate a Gaussian (RBF) kernel.

This procedure was refined by \citet{LeSarSmo13} by
approximating the sampling step with a combination of matrices that admit fast
computation.  They show that one may compute \emph{Fastfood} approximate kernel expansions via
\begin{align}
  \label{eq:Fastfood}
  \tilde{k}(x,x') \propto \frac{1}{m} \sum_{j=1}^m \phi_j(x)
  \phi^{*}_j(x')
                    \text{ where }
  \phi_j(x)  = \exp \rbr{i [S H G \Pi H B x]_j}.
\end{align}
The random matrices $S, H, G, \Pi, B$ are chosen such as to provide a
sufficient degree of randomness while also allowing for efficient computation.
\begin{description}
\item[$B$ Binary decorrelation] The entries $B_{ii}$ of this diagonal matrix
  are drawn uniformly from $\cbr{\pm 1}$. This ensures that the data
  have zero mean in expectation over all matrices $B$.
\item[$H$ Hadamard matrix] It is defined recursively via
  \begin{align*}
    H_{1} := \mymatrix{r}{1}
    \text{ and }
    H_{2d} := \mymatrix{rr}{H_d & H_d \\ H_d & -H_d}
    \text{ hence }  H_{2d} \mymatrix{l}{x \\ x'} =
    \mymatrix{l}{H_d [x+x'] \\ H_d [x - x']}.
  \end{align*}
  The recursion shows that the dense matrix $H_d$ admits fast multiplication in
  $O(d \log d)$ time, i.e.\ as efficiently as the FFT allows.
\item[$\Pi$ Permutation matrix] This decorrelates the eigensystems of
  subsequent Hadamard matrices. Generating such a random permutation
  (and executing it) can be achieved by reservoir sampling, which
  amounts to $n$ in-place pairwise swaps. It ensures that the spaces
  of both permutation matrices are effectively uncorrelated.
\item[$G$ Gaussian matrix] It is a diagonal matrix with Gaussian entries drawn
  iid via $G_{ii} \sim \Ncal(0, 1)$. The result of using it is that
  each of the rows of $HG\Pi HB$ consist of iid Gaussian random
  variables. Note, though, that the rows of this matrix are not quite
  independent.
\item[$S$ Scaling matrix] This diagonal matrix encodes the spectral properties
  of the associated kernel. Consider $\rho(\omega)$ of \eq{eq:transinv}. There
  we draw $\omega$ from the spherically symmetric distribution defined by
  $\rho(\omega)$ and use its length to rescale $S_{ii}$ via
  \begin{align}
    \nonumber
    S_{ii} = \nbr{\omega_i} \nbr{G}^{-1}_\mathrm{Frob}
  \end{align}
\end{description}
It is straightforward to change kernels, for example, by adjusting $S$. Moreover, all the computational benefits of decomposing terms via
\eq{eq:Fastfood} remain even after adjusting $S$.  Therefore we can customize kernels for the problem at hand
rather than applying a generic kernel, without incurring additional computational
expenses.

\section{\`{A} la Carte}
\label{sec:alacarte}

In keeping with the culinary metaphor of Fastfood, we now introduce a flexible
and efficient approach to kernel learning \emph{\`{a} la carte}. That
is, we will adjust the spectrum of a kernel in such a way as to allow for a
wide range of translation-invariant kernels. Note that unlike previous
approaches, this can be accomplished without any additional cost since
these kernels only differ in terms of their choice of scaling.

In Random Kitchen Sinks and Fastfood, the frequencies $\omega$ are sampled from
the spectral density $\rho(\omega)$.  One could instead learn the frequencies
$\omega$ using a kernel learning objective function.  Moreover, with enough
spectral frequencies, such an approach could learn any stationary (translation
invariant) kernel.  This is because each spectral frequency corresponds to a
point mass on the spectral density $\rho(\omega)$ in \eqref{eq:transinv}, and
point masses can model any density function.

However, since there are as many spectral frequencies as there are basis
functions, individually optimizing over all the frequencies $\omega$ can still
be computationally expensive, and susceptible to over-fitting and many
undesirable local optima. In particular, we want to enforce smoothness over the
\emph{spectral distribution}.  We therefore also propose to learn the scales,
spread, and locations of \emph{groups} of spectral frequencies, in a procedure
that modifies the expansion \eq{eq:Fastfood} for fast kernel learning.  This
procedure results in efficient, expressive, and lightly parametrized models.

In sections \ref{sec:learnkernel} and \ref{sec:marglike} we describe a procedure for learning the free
parameters of these models, assuming we already have a Fastfood expansion.
Next we introduce four new models under this efficient framework -- a Gaussian
spectral mixture model in section \ref{sec:spectralmix}, a piecewise linear
radial model in section \ref{sec:piecewise}, and models which learn the
scaling ($S$), Gaussian ($G$), and binary decorrelation ($B$) matrices in
Fastfood in section \ref{sec:optfast}.
\subsection{Learning the Kernel}
\label{sec:learnkernel}

We use a Gaussian process (GP) formalism for kernel learning.  For an
introduction to Gaussian processes, see \citet{RasWil06}, for example.
Here we assume we
have an efficient Fastfood basis function expansion for kernels of interest; in
the next sections we derive such expansions.

For clarity, we focus on regression, but we note our methods can be used 
for classification and non-Gaussian likelihoods without additional methodological
innovation.  When using Gaussian processes for classification, for example, one 
could use standard approximate Bayesian inference to represent an approximate
marginal likelihood \citep{RasWil06}.  A primary goal of this paper is to demonstrate how Fastfood can be extended to
learn a kernel, independently of a \emph{specific} kernel learning objective.
However, the marginal likelihood of a Gaussian process provides a general purpose
probabilistic framework for kernel learning, particularly suited to training
highly expressive kernels \citep{Wilson14}.  Note that there are many other
choices for kernel learning objectives.  For
instance, \citet{OngSmoWil03} provide a rather encyclopedic list of
alternatives.

Denote by $\Xcal$ an index set with $X := \cbr{x_1, \ldots x_n}$ drawn
from it. We assume that the observations $y$ are given by
\begin{align}
  \label{eq:gpdef}
  y = f + \epsilon \,,
  \text{ where } \epsilon \sim \Ncal\rbr{0, \sigma^2}.
\end{align}
Here $f$ is drawn from a Gaussian process $\gp(0,
k_{\gamma})$.
Eq.~\eqref{eq:gpdef} means that any \emph{finite dimensional realization}
$f \in \RR^{n}$ is drawn from a normal distribution $\Ncal(0, K)$, where
$K_{ij} = k(x_i, x_j)$ denotes the associated kernel matrix. This also means
that $y \sim \Ncal(0, K + \sigma^2 I)$, since the additive noise $\epsilon$ is
drawn iid for all $x_i$.
For finite-dimensional feature spaces we can equivalently use the
representation of \citet{Williams98}:
\begin{align*}
  f(x)  = \inner{w}{\phi(x)} \,, & \text{ where } w  \sim \Ncal(0, \sigma^2
  I) \\
  \text{hence } f \sim \gp(0, k) \,, & \text{ where } k(x,x') = \sigma^2 \inner{\phi(x)}{\phi(x')}
\end{align*}
The kernel of the Gaussian process is parametrized by $\gamma$.  Learning the kernel
therefore involves learning $\gamma$ and $\sigma^2$ from the data, or
equivalently, inferring the structure of the feature map $\phi(x)$.

Our working assumption is that $k_\gamma$ corresponds to a
$Q$-component mixture model of kernels $k_q$ with associated weights
$v_q^2$. Moreover we assume that we have access to a
Fastfood expansion $\phi_q(x)$ for each of the components into $m$
terms. This leads to
\begin{align*}
  f(x) & = \sum_{q=1}^Q \sum_{j=1}^m w_{qj} \phi_{qj}(x|\theta_q)
  \text{ where } w_{qj} \sim \Ncal\rbr{0, m^{-1} v_q^2} \\
  k(x,x') & = \sum_{q=1}^Q \frac{v_q^2}{m} \sum_{j=1}^m
  \phi_{qj}(x|\theta_q) \phi_{qj}(x'|\theta_q)
\end{align*}
This kernel is parametrized by $\gamma = \cbr{v_q, \theta_q}$. Here $v_q$
are mixture weights and $\theta_q$ are parameters of the (non-linear)
basis functions $\phi_{qj}$.

\subsection{Marginal Likelihood}
\label{sec:marglike}

We can marginalise the Gaussian process governing $f$ by
integrating away the $w_{qj}$ variables above to express the
marginal likelihood of the data solely in terms of the kernel hyperparameters
$v, \theta$ and noise variance $\sigma^2$.

Denote by $\Phi_{\theta} \in \RR^{Qm \times n}$ the design matrix, as
parametrized by $\theta$, from evaluating the functions
$\phi_{qj}(x|\theta_q)$ on $X$. Moreover, denote by $V \in \RR^{Q m
  \times Q m}$ the diagonal
scaling matrix obtained from $v$ via
\begin{align}
  \label{eq:Theta}
  V := m^{-1} \mathrm{diag}(v_1, \ldots, v_1, \ldots,
  v_Q, \ldots v_Q).
\end{align}
Since $\epsilon$ and $f$ are independent, their covariances are
additive.  For $n$ training datapoints $y$, indexed by $X$, we
therefore obtain the marginal likelihood
\begin{align}
  \label{eq:integratedout}
  {y}|X, v, \theta, \sigma^2 & \sim \Ncal(0, \Phi_{\theta}^\top V \Phi_{\theta} + \sigma^2 I)
\end{align}
and hence the negative log marginal likelihood is
\begin{align}
  \nonumber
  -\log p(y|X, \gamma, \sigma^2) = & \frac{n}{2} \log 2\pi + \frac{1}{2} \log
  \abr{\Phi^\top_{\theta} V \Phi_{\theta} + \sigma^2 I}
  +\frac{1}{2} {y}^\top \sbr{\Phi_{\theta}^\top V \Phi_{\theta} + \sigma^2
    I}^{-1} {y}
  \label{eq:integratedout-log}
\end{align}
To learn the kernel $k$ we minimize the negative log marginal
likelihood of \eq{eq:integratedout-log}
with respect to $v, \theta$ and $\sigma^2$.
Similarly, the predictive distribution at a test input $\bar{x}$ can be
evaluated using
\begin{align}
  \bar{y}|\bar{x},X,y,v,\theta,\sigma^2
  & \sim \Ncal(\bar\mu,{\bar\sigma}^2) \\
  \nonumber
  \text{where }
  \bar\mu & = k(\bar{x})^\top \sbr{\Phi^\top V \Phi + \sigma^2 I}^{-1}
  y \text{ and }
  {\bar\sigma}^2 = \sigma_n^2 + \bm{k}(\bar{x})^\top \sbr{\Phi^\top
    V \Phi + \sigma^2 I}^{-1} \bm{k}(\bar{x}).
\end{align}
Here $\bm{k}(\bar{x}) := \rbr{k(\bar{x}, x_1), \ldots, k(\bar{x}, x_n)}^{\top}$
denotes the vector of cross covariances between the test point $x$ and
and the $n$ training points in $X$.
On closer inspection we note that these expressions can be simplified
greatly in terms of $\Phi$ and $\phi_{qj}(\bar{x})$ since
\begin{align*}
  k(\bar{x}) = \phi(\bar{x})^\top V \Phi \text{ and hence }
  \bar\mu = \phi(\bar{x})^\top \beta \text{ for }
  \beta = V \sbr{\Phi^\top V \Phi + \sigma^2 I}^{-1} {y}
\end{align*}
with an analogous expression for $\bar\sigma^2$. More importantly,
instead of solving the problem in
terms of the kernel matrix we can perform inference in terms of
$\beta$ directly. This has immediate benefits:
\begin{itemize}
\item Storing the solution only requires $O(Q m)$ parameters
  regardless of $X$, provided
  that $\phi(x)$ can be stored and computed efficiently, e.g.\ by
  Fastfood.
\item Computation of the predictive variance is equally efficient:
  $\Phi^\top V \Phi$ has at most rank $Q m$, hence the
  evaluation of $\bar\sigma^2$ can be accomplished via the
  Sherman-Morrison-Woodbury formula, thus requiring only $\mathcal{O}(Q^2 m^2 n)$
  computations. Moreover, randomized low-rank approximations of
  \begin{align*}
    V^{\frac{1}{2}} \Phi \sbr{\Phi^\top V \Phi +
    \sigma_n^2 I}^{-1} \Phi^\top V^{\frac{1}{2}},
  \end{align*}
  using
  e.g.\ randomized projections, as proposed by \citet{HalMarTro09}, allow for even more
  efficient computation.
\end{itemize}
Overall, standard Gaussian process kernel representations \citep{RasWil06} require
$\mathcal{O}(n^3)$ computations and $\mathcal{O}(n^2)$ memory.  Therefore,
when using a Gaussian process kernel learning formalism,
the expansion in this section, using $\Phi$, is computationally preferable
whenever $Qm < n$.

\subsection{Gaussian Spectral Mixture Models}
\label{sec:spectralmix}

For the Gaussian Spectral Mixture kernels of \citet{WilAda13}, translation
invariance holds, yet rotation invariance is violated:
the kernels satisfy $k(x,x') = k(x+\delta, x'+\delta)$ for all
$\delta \in \RR^d$; however, in general rotations $U \in
\mathrm{SO}(d)$ do not leave $k$ invariant, i.e.\
$k(x,x') \neq k(Ux, Ux')$. These kernels have the following
explicit representation in terms of their Fourier transform $F[k]$
\begin{align*}
  F[k](\omega)  = \sum_q \frac{v^2_q}{2} \sbr{\chi\rbr{\omega, \mu_q, \Sigma_q} +
    \chi\rbr{-\omega, \mu_q, \Sigma_q}}
                 \text{ where }
  \chi(\omega, \mu, \Sigma)  = \frac{e^{-\frac{1}{2} (\mu - \omega)^\top \Sigma^{-1} (\mu - \omega)}}{(2 \pi)^{\frac{d}{2}} |\Sigma|^{\frac{1}{2}} }
\end{align*}
In other words, rather than choosing a spherically symmetric representation
$\rho(\omega)$ as typical for \eq{eq:transinv}, \citet{WilAda13} pick a mixture
of Gaussians with mean frequency $\mu_q$ and variance $\Sigma_q$ that satisfy
the symmetry condition $\rho(\omega)= \rho(-\omega)$ but not rotation
invariance. By the linearity of the Fourier transform, we can apply the inverse
transform $F^{-1}$ component-wise to obtain
\begin{align}
  \label{eq:inverse-mix}
  k(x-x')  = &\sum_q v^2_q \frac{\abr{\Sigma_q}^\frac{1}{2}}{(2
    \pi)^{\frac{d}{2}}}
  \exp\rbr{-\frac{1}{2} \nbr{\Sigma_q^{\frac{1}{2}} (x-x')}^2} \cos \inner{x-x'}{\mu_q}
\end{align}
\begin{lemma}[Universal Basis]
  The expansion \eq{eq:inverse-mix} can approximate any
  translation-invariant kernel by approximating its spectral density.
\end{lemma}
\begin{proof}
  This follows since mixtures of Gaussians are
  universal approximators for densities \citep{Silverman86}, as is well
  known in the kernel-density estimation literature. By the
  Fourier-Plancherel theorem, approximation in Fourier domain amounts to
  approximation in the original domain, hence the result applies to
  the kernel.
\end{proof}
%
Note that the expression in \eqref{eq:inverse-mix} is \emph{not} directly amenable to the
fast expansions provided by Fastfood since the distributions are
shifted. However, a small modification allows us to efficiently
compute kernels of the form of \eq{eq:inverse-mix}. The key insight is
that shifts in Fourier space by $\pm\mu_q$ are accomplished by
multiplication by $\exp\rbr{\pm i \inner{\mu_q}{x}}$. Here the inner
product can be precomputed, which costs only $O(d)$ operations. Moreover,
multiplications by $\Sigma_q^{-\frac{1}{2}}$ induce multiplication by
$\Sigma_q^{\frac{1}{2}}$ in the original domain, which can be
accomplished as \emph{preprocessing}. For diagonal $\Sigma_q$ the
cost is $O(d)$.

In order to preserve translation invariance we compute a symmetrized
set of features. We have the following algorithm (we assume diagonal
$\Sigma_q$ --- otherwise simply precompute and scale $x$):

\begin{algorithmic}
  \STATE {\bfseries Preprocessing}
  {\bfseries --- Input} $m, \cbr{(\Sigma_q, \mu_q)}$
  \STATE {\bfseries for each $q$} generate random matrices $S_q, G_q,
  B_q, \Pi_q$
  \STATE Combine group scaling $B_q \leftarrow B_q \Sigma_q^{\frac{1}{2}}$ \\[2mm]
  \STATE {\bfseries Feature Computation}
  {\bfseries --- Input} $S, G, B, \Pi, \mu, \Sigma$
  \FOR{$q=1$ {\bfseries to} $Q$}
  \STATE $\zeta \leftarrow \inner{\mu_q}{x}$ (offset)
  \STATE $\xi \leftarrow [S_q H G_q \Pi_q H B_q x]$ (Fastfood product)
  \STATE Compute features %
  \begin{align*}
    \phi_{q \cdot 1} \leftarrow\sin (\xi + \zeta)
    \text{ and }
    \phi_{q \cdot 2} \leftarrow \cos (\xi + \zeta)
    \text{ and }
    \phi_{q \cdot 3} \leftarrow\sin (\xi - \zeta)
    \text{ and }
    \phi_{q \cdot 4} \leftarrow \cos (\xi + \zeta)
  \end{align*}
  \ENDFOR
\end{algorithmic}

To learn the kernel we learn the weights $v_q$, dispersion
$\Sigma_q$ and locations $\mu_q$ of spectral frequencies
via marginal likelihood optimization, as described in
section \ref{sec:marglike}.  This results in a kernel
learning approach which is similar in flexibility to individually learning
all $md$ spectral frequencies and is less prone to over-fitting and local
optima. In practice, this can mean optimizing over about $10$
free parameters instead of $10^4$ free parameters, with improved predictive
performance and efficiency.  See section \ref{sec:experiments} for more detail.

\subsection{Piecewise Linear Radial Kernel}
\label{sec:piecewise}

In some cases the freedom afforded by a mixture of Gaussians in
frequency space may be more than what is needed. In particular, there
exist many cases where we \emph{want} to retain invariance under
rotations while simultaneously being able to adjust the spectrum
according to the data at hand.   For this purpose we introduce
a novel piecewise linear radial kernel.

Recall \eq{eq:transinv} governs the regularization
properties of $k$. We require $\rho(\omega) = \rho(\nbr{\omega}) := \rho(r)$ for
rotation invariance. For instance, for the Gaussian RBF kernel we have
\begin{align}
  \rho(\nbr{\omega}_2) \propto \nbr{\omega}_2^{d-1} \exp\rbr{\textstyle-\frac{\nbr{\omega}^2_2}{2}}.
\end{align}
For high dimensional inputs, the RBF kernel suffers from a concentration of
measure problem \citep{LeSarSmo13}, where samples are tightly concentrated at
the maximum of $\rho(r)$, $r = \sqrt{d-1}$.
A fix is relatively easy, since we are at liberty to pick
any nonnegative $\rho$ in designing kernels. This procedure is flexible but
leads to intractable integrals: the Hankel transform of $\rho$, i.e.\
the radial part of the Fourier transform, needs to be analytic if we
want to compute $k$ in closed form.

\begin{figure}[t]
  \centering
  \begin{minipage}{0.49\textwidth}
    \centering
    $\displaystyle\rho_i(r)$ \\[-5mm]
    \includegraphics[width=\textwidth]{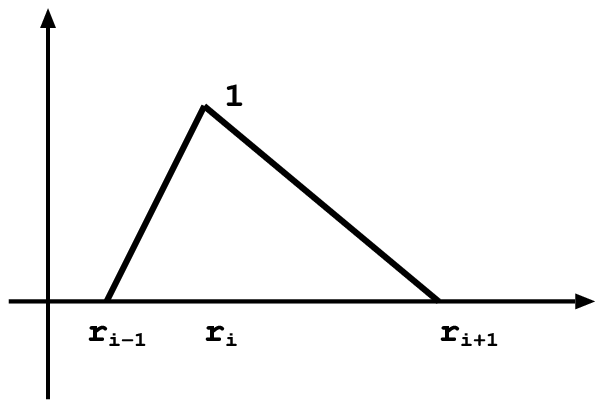}
  \end{minipage}
  \begin{minipage}{0.49\textwidth}
    \centering
    $\displaystyle\rho(r) = \sum_i \alpha_i \rho_i(r)$  \\[-5mm]
    \includegraphics[width=\textwidth]{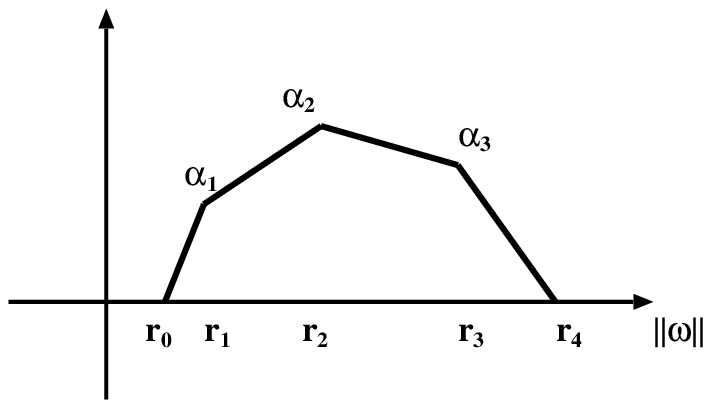}
  \end{minipage}
  \caption{Piecewise linear functions. Left: single basis
      function. Right: linear combination of three functions.
      Additional degrees of freedom are fixed by $\rho(r_0) =
      \rho(r_4) = 0$.
      \label{fig:piecewiseboth}}
\end{figure}

However, if we remain in the Fourier domain, we can use $\rho(r)$ and
sample \emph{directly} from it. This strategy kills two birds with one
stone: we do not need to compute the inverse Fourier transform and
we have a readily available sampling distribution at our
disposal for the Fastfood expansion coefficients $S_{ii}$. All that is
needed is to find an efficient parametrization of $\rho(r)$.

We begin by providing an explicit expression for piecewise linear
functions $\rho_i$ such that $\rho_i(r_j) = \delta_{ij}$
with discontinuities only at
$r_{i-1}, r_i$ and $r_{i+1}$. In other words, $\rho(r)$ is a `hat' function
with its mode at $r_i$ and range $[r_{i-1}, r_{i+1}]$. It is
parametrized as
\begin{align}
  \nonumber
  \rho_i(r) := \max\rbr{0, \min\rbr{1, \frac{r-r_{i-1}}{r_i
        - r_{i-1}}, \frac{r_{i}-r}{r_{i+1} - r_{i}}}}
\end{align}
By construction each basis function is piecewise linear with
$\rho_i(r_j) = \delta_{ij}$ and moreover $\rho_i(r) \geq 0$ for all
$r$.
\begin{lemma}
  Denote by $\cbr{r_0, \ldots, r_n}$ a sequence of locations with $r_i
  > r_{i-1}$ and $r_0 = 0$. Moreover, let $\rho(r) := \sum_i \alpha_i
  \rho_i(r)$. Then $\rho(r) \geq 0$ for all $r$ if and only if $\alpha_i
  \geq 0$ for all $i$. Moreover, $\rho(r)$ parametrizes all piecewise
  linear functions with discontinuities at $r_i$.
\end{lemma}
Now that we have a parametrization we only need to discuss how to draw
$\omega$ from $\rho(\nbr{\omega}) = \rho(r)$. We have several strategies at our disposal:
\begin{itemize}
\item $\rho(r)$ can be normalized explicitly via
  $$\bar\rho := \int_{0}^\infty \rho(r) dr = \sum_i
  \frac{\alpha_i}{2(r_{i+1} - r_{i-1})}$$
  Since each segment $\rho_i$ occurs with probability $\alpha_i/(2
  \bar\rho (r_{i+1} - r_{i-1})$ we first sample the segment and then
  sample from $\rho_i$ explicitly by inverting the associated
  cumulative distribution function (it is piecewise quadratic).
\item Note that sampling can induce considerable variance in the
  choice of locations. An alternative is to invert the cumulative
  distribution function and pick $m$
  locations equidistantly at locations $\frac{i}{m} + \xi$ where $\xi
  \sim U[0, 1/m]$. This approach is commonly used in particle
  filtering \citep{DouFreGor01}. It is equally cheap yet substantially
  reduces the variance when sampling, hence we choose this strategy.
\end{itemize}
The basis functions are computed as follows:

\begin{algorithmic}
  \STATE {\bfseries Preprocessing}($m, \cbr{(\{\alpha_i\}_{i=1}^n,\{ r_i\}_{i=0}^{n+1},  \Sigma)}$)
  \STATE Generate random matrices $G, B, \Pi$
  \STATE Update scaling $B \leftarrow B \Sigma^{\frac{1}{2}}$
  \STATE Sample $S$ from $\rho(\nbr{\omega})$ as above  \\[2mm]
  \STATE {\bfseries Feature Computation}($S, G, B, \Pi$)
  \begin{align*}
    \phi_1 \leftarrow \cos([S H G \Pi H B x]) \text{ and }
    \phi_2 \leftarrow  \sin([S H G \Pi H B x])
    \end{align*}
\end{algorithmic}
The rescaling matrix $\Sigma_q$ is introduced to incorporate automatic relevance
determination into the model. Like with the Gaussian spectral mixture model, we can
use a mixture of piecewise linear radial kernels to approximate any radial kernel.
Supposing there are $Q$ components of the piecewise linear $\rho_q(r)$  function, we
can repeat the proposed algorithm $Q$ times to generate all the required basis functions.

\subsection{Fastfood Kernels}
\label{sec:optfast}

The efficiency of Fastfood is partly obtained by approximating Gaussian random
matrices with a product of matrices described in section \ref{sec:Fastfood}.
Here we propose several expressive and efficient kernel learning algorithms
obtained by optimizing the marginal likelihood of the data in Eq.~\eqref{eq:integratedout-log}
with respect to these matrices:
\begin{description}
\item[FSARD] The scaling matrix $S$ represents the spectral properties of the
  associated kernel. For the RBF kernel, $S$ is sampled from a chi-squared
  distribution. We can simply change the kernel by adjusting $S$. By varying
  $S$, we can approximate any radial kernel. We learn the diagonal matrix
  $S$ via marginal
  likelihood optimization.  We combine this procedure with \emph{Automatic
    Relevance Determination} of \citet{Neal98} -- learning the scale
  of the input space -- to obtain the {\bf FSARD} kernel.
\item[FSGBARD] We can further generalize {\bf FSARD} by additionally
  optimizing marginal likelihood with respect to the diagional matrices $G$ and $B$ in
  Fastfood to represent a wider class of kernels.
\end{description}
In both {\bf FSARD} and {\bf FSGBARD} the Hadamard matrix $H$ is
retained, preserving all the computational benefits of Fastfood. That
is, we only modify the scaling matrices while keeping the main
computational drivers such as the fast matrix multiplication and the
Fourier basis unchanged.

\section{Experiments}
\label{sec:experiments}

We evaluate the proposed kernel learning algorithms
on many regression problems from the UCI
repository. We show that the proposed methods are
flexible, scalable, and applicable
to a large and diverse collection of data, of varying sizes and
properties. In particular, we demonstrate scaling to more than 2
million datapoints (in general,
Gaussian processes are intractable beyond $10^4$ datapoints); secondly,
the proposed algorithms significantly outperform
standard exact kernel methods, and with only a few hyperparameters are
even competitive with alternative methods that involve training
orders of magnitude more hyperparameters.\footnote{GM, PWL, FSARD, and FSGBARD are novel
contributions of this paper, while RBF and ARD are popular alternatives,
and SSGPR is a recently proposed state of the art kernel learning approach.}
The results are shown in Table \ref{table: testresults}.
All experiments are performed on an Intel Xeon E5620 PC, operating at
2.4GHz with 32GB RAM.

\subsection{Experimental Details}
To learn the parameters of the kernels, we optimize over the marginal
likelihood objective function described in Section 3.1, using
LBFGS.\footnote{\url{http://www.di.ens.fr/\~mschmidt/Software/minFunc.html}}

The datasets are divided into three groups: \texttt{SMALL} $n\leq 2000$ and
\texttt{MEDIUM} $2,000 < n \leq 100,000$ and \texttt{LARGE}
$100,000 < n\leq 2,000,000$. All methods -- RBF, ARD, FSARD, GM, PWL and SSGPR
-- are tested on each grouping. For \texttt{SMALL} data, we use an exact RBF
and ARD kernel. All the datasets are divided into $10$ partitions. Every time,
we pick one partition as test data and train all methods on the remaining $1$
partitions. The reported result is based on the averaged RMSE of $10$
partitions.

\paragraph{Methods}
\begin{description}
\item[RBF and ARD] 
  The RBF kernel has 
  the form $k(x,x') = a^2 \exp \left(-0.5 ||x-x'||^2 / \ell^2\right)$, where $a$ and $\ell$ are 
  signal standard deviation and length-scale (rescale) hyperparameters.  The
  ARD kernel has the form $k(x,x') = a^2 \exp \left( - 0.5 \sum_{j=1}^d (x_j - x_j')^2 / \ell_j^2 \right)$.
  ARD kernels use Automatic Relevance Determination
  \citep{Neal98} to adjust the scales of each input dimension individually.  
  On smaller datasets, with fewer than $n = 2000$ training
  examples, where exact methods are tractable, we use exact Gaussian RBF and
  ARD kernels with hyperparameters learned via marginal likelihood
  optimization.  Since these
  exact methods become intractable on larger datasets, we use Fastfood basis
  function expansions of these kernels for $n > 2000$.  
\item[GM] For Gaussian Mixtures we compute a mixture of Gaussians in frequency
  domain, as described in section~\ref{sec:spectralmix}. As before,
  optimization is carried out with regard to marginal likelihood.
\item[PWL] For rotation invariance, we use the novel piecewise
  linear radial kernels described in section~\ref{sec:piecewise}.  PWL has a
  simple and scalable parametrization in terms of the radius of the spectrum.  
\item[SSGPR] Sparse Spectrum Gaussian Process Regression is a kitchen sinks
  (RKS) based model which individually optimizes the locations of \emph{all}
  spectral frequencies \citep{Lazaroetal10}.  We note that SSGPR is heavily
  parametrized. Also note that SSGPR is a special case of the proposed GM model
  if for GM the number of components $Q=m$, and we set all bandwidths to
  $0$ and weigh all terms equally.
\item[FSARD and FSGBARD] As described in section~\ref{sec:optfast}, these
  methods respectively learn the $S$ and $S, G, B$ matrices in the Fastfood
  representation of section~\ref{sec:Fastfood}, through marginal likelihood
  optimisation (section \ref{sec:marglike}).
\end{description}

\paragraph{Initialization}
\begin{description}
\item[RBF] We randomly pick $max(2000, n/5)$ pairs of data and compute the
  distance of these pairs. These distances are sorted and we pick the
  $[0.1:0.2:0.9]$ quantiles as length-scale (aka rescale) initializations.
  We run $20$ optimization iterations starting with
  these initializations and then pick the one with the minimum negative
  log marginal likelihood, and then continue to optimize for $150$ iterations. We
  initialize the signal and noise standard deviations as $std(y)$ and
  $std(y)/10$, respectively, where $y$ is the data vector.
\item[ARD] For each dimension of the input, $X_j$, we initialize each length-scale
  parameter as $\ell_j = u(max(X_j) - min(X_j))$, where
  $u \sim \text{Uniform} [0.4, 0.8]$.  We pick the best initialization from
  $10$ random restarts of a $20$ iteration optimization run.
  We multiply the scale by $\sqrt{d}$, the total number of input dimensions.
\item[FSARD] We use the same technique as in ARD to initialize the rescale
  parameters.  We set the $S$ matrix to be $||w||$, where $w$ is a $d$
  dimensional random vector with standard Gaussian distribution.
\item[FSGBARD] We initialize $S$ as in FSARD. $B$ is
  drawn uniformly from $\cbr{\pm 1}$ and $G$ from draws of a standard Gaussian
  distribution.
\item[GM] We use a Gaussian distribution with diagonal covariance matrices
  to model the spectral density $p(\omega)$. Assuming there are $Q$ mixture components, the scale of
  each component is initialized to be $std(y)/Q$. We reuse the same technique
  to initialize the rescale matrices as that of ARD. We initialize the shift $\mu$ to
  be close to $0$.
\item[PWL] We make use of a special case of a piecewise linear function, the
  hat function, in the experiments. The hat function is parameterized by $\mu$
  and $\sigma$.  $\sigma$ controls the width of the hat function and $\mu$
  control the distance of the hat function from the origin. We also incorporate
  ARD in the kernel, and use the same initialization techniques. For the RBF
  kernel we compute the distance of random pairs and sort these distances. Then
  we get a distance sample $\lambda$ with a uniform random quantile within
  $[0.2,0.8]$.  $\lambda$ is like the bandwidth of an RBF kernel. Then
  $\sigma = 2/\lambda$ and $\mu = \max\{\sqrt{d-1}- 2, 0.01\}/\lambda$. We make
  use of this technique because for RBF kernel, the maximum points is at
  $\sqrt{d-1}/\lambda$ and the width of the radial distribution is about
  $2/\lambda$.
\item[SSGPR] For the rescale parameters, we follow the same procedure as for the ARD kernel.
We initialize the projection matrix as a random Gaussian matrix.
\end{description}

\begin{figure}[htb]
  \centering
  \includegraphics[width=0.49\textwidth]{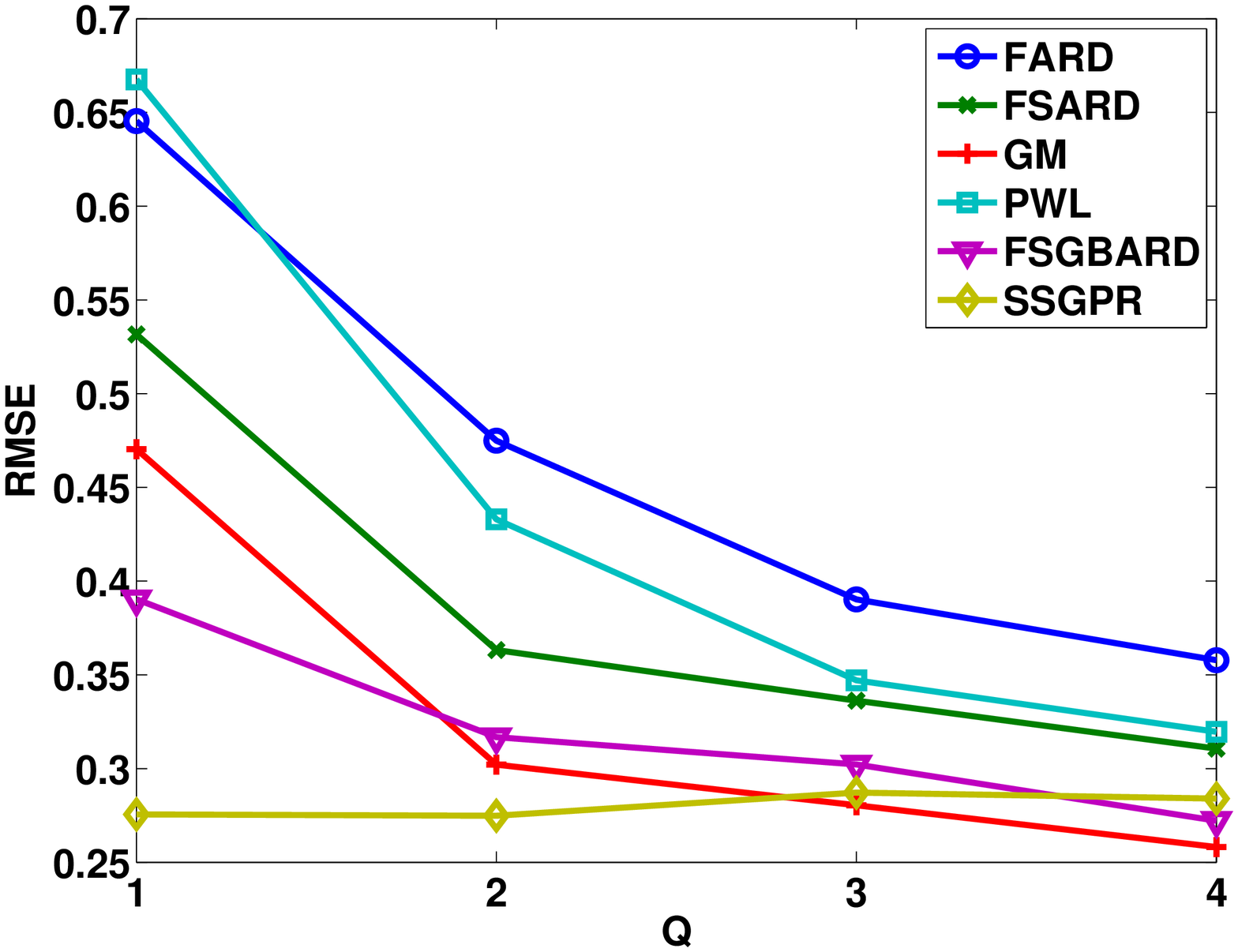}
  \includegraphics[width=0.49\textwidth]{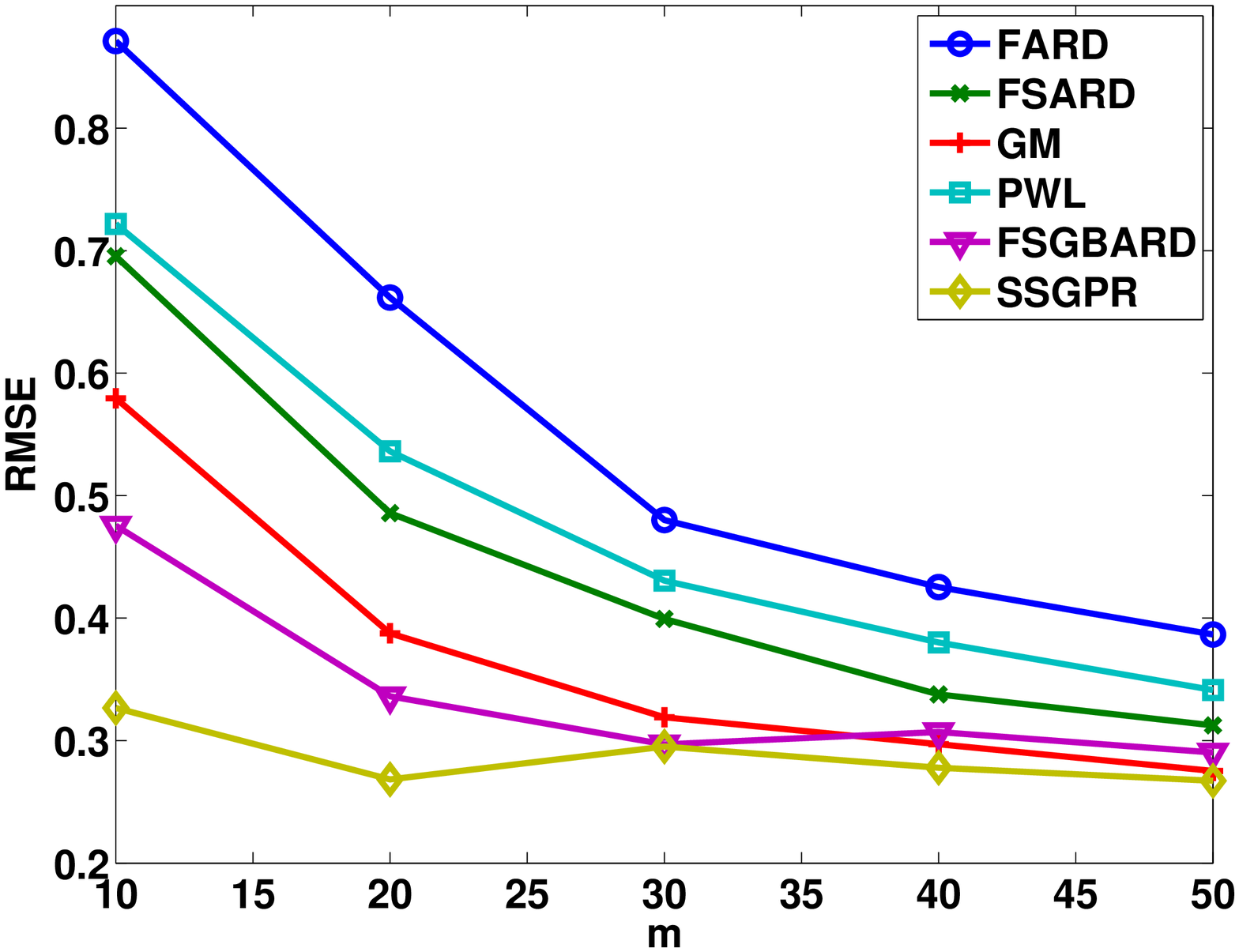}
  \caption{\label{fig:qbasis}%
    We analyze how the accuracy depends on the number of
    clusters $Q$ (left) and the number of basis functions $m$ (right).
    More specifically, for variable $Q$
    the number of basis functions per group $m$ is fixed to $32$. For variable
    $m$ the number of clusters $Q$ is fixed to $2$.  FRBF and FARD are Fastfood
    expansions of RBF and ARD kernels, respectively.}
\end{figure}
\begin{figure}[htb]
  \centering
  \includegraphics[width=0.49\textwidth]{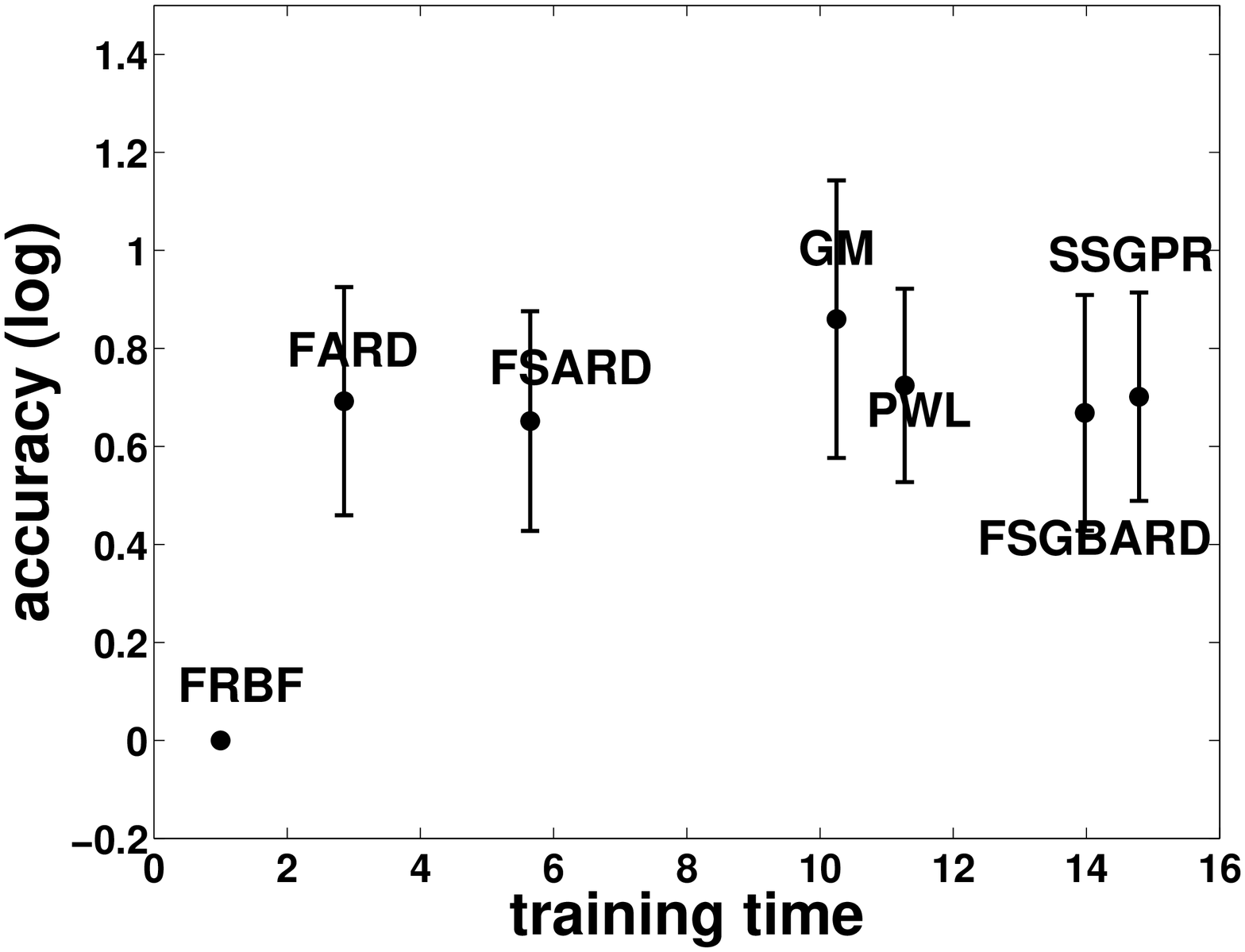}
  \includegraphics[width=0.49\textwidth]{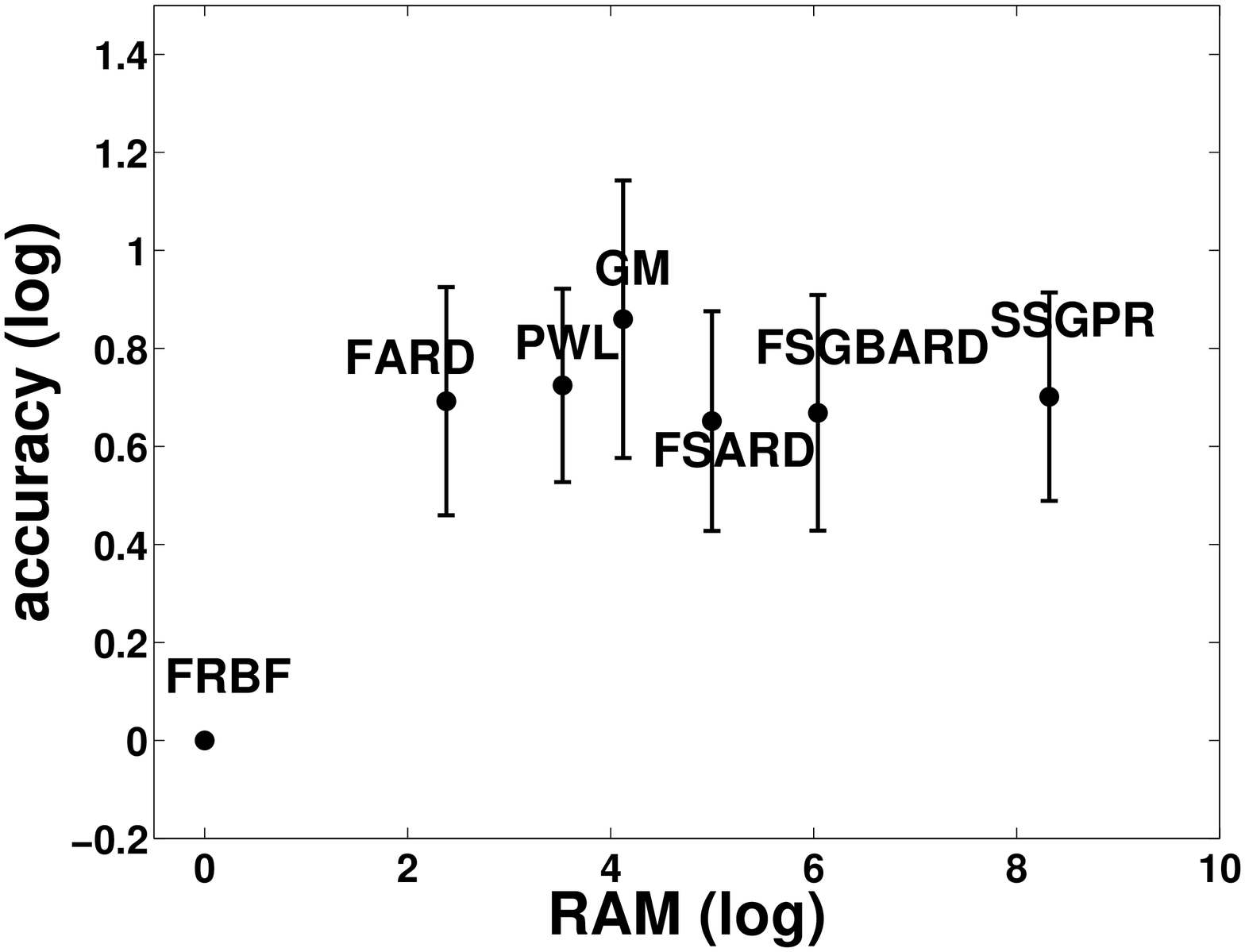}
  \caption{\label{fig:accutimelog}%
    We compare all methods in terms of accuracy, training time (left) and
    memory (right). To make the methods comparable we compute the
    accuracy score of a given method as the improvement relative to
    the baseline (FRBF). That is, we compute
    $\text{accuracy}_{\text{method}}=\text{RMSE}_{\text{FRBF}}/\text{RMSE}_{\text{method}}$.
    By construction, FRBF has an accuracy score of $1$, and
    larger values of accuracy correspond to better algorithms.
    For runtime and memory we take the reciprocal of the analogous metric, so
    that a lower score corresponds to better performance.  For instance,
    $\text{time}_{\text{method}}=\text{walltime}_{\text{method}}/\text{walltime}_{\text{FRBF}}$.
    $\log$ denotes an average of the log scores, across all
    datasets. 
}
\end{figure}

\begin{figure}[htb]
  \centering
  \includegraphics[width=0.49\textwidth]{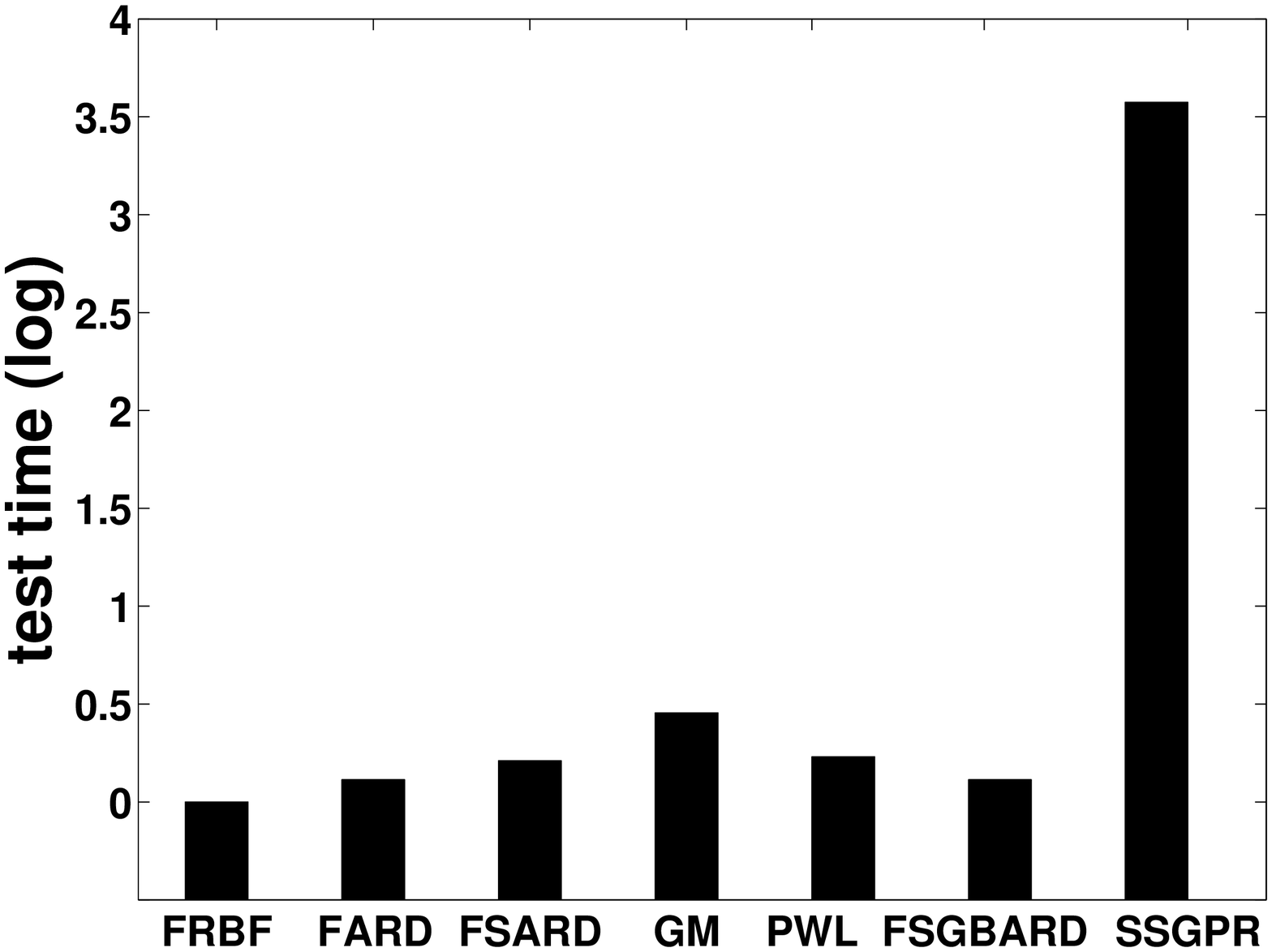}
  \includegraphics[width=0.49\textwidth]{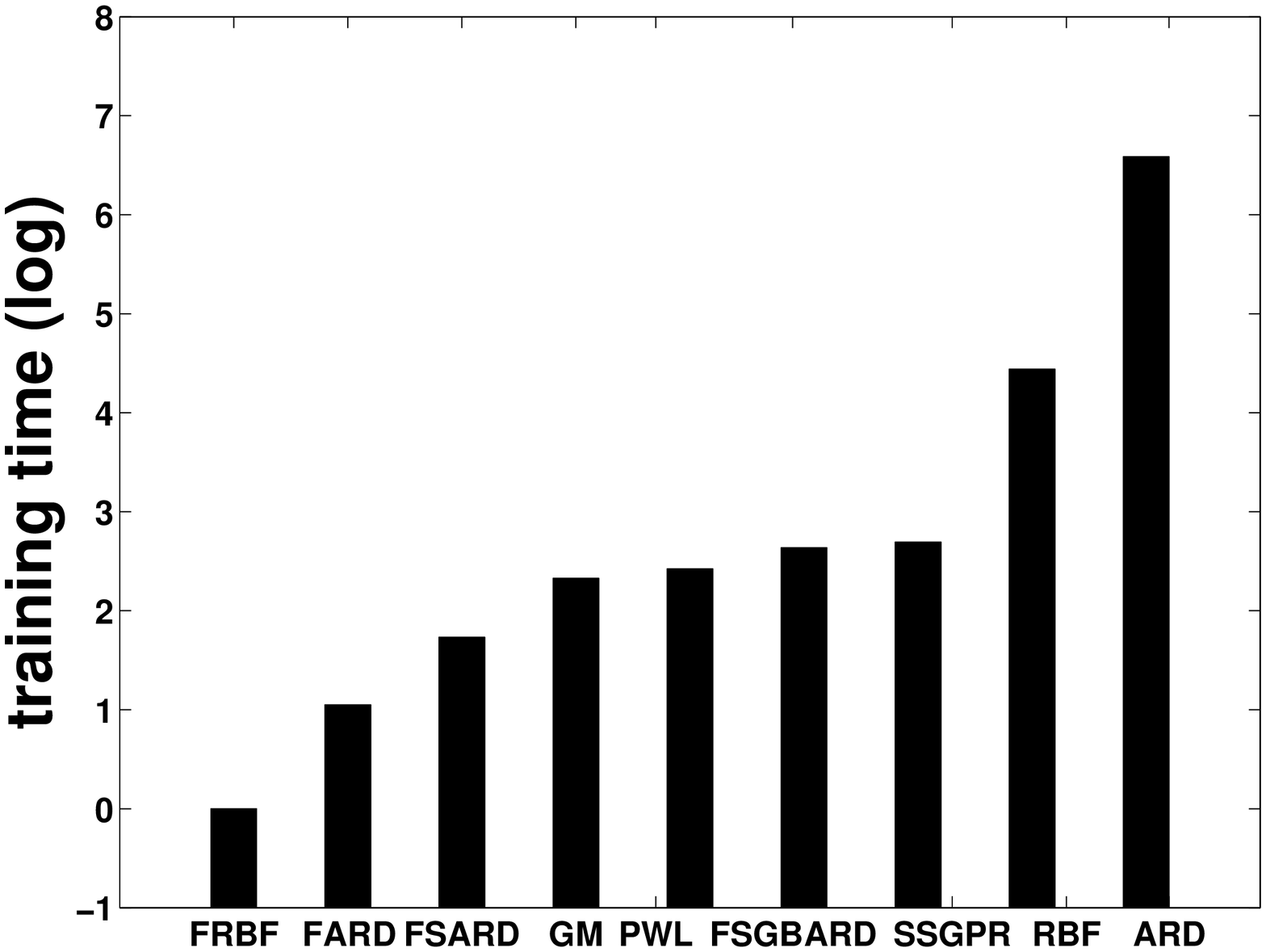}
  \caption{\label{fig:testtime}%
    Test time (left) and training time (right), as a function of the
    method used. This also includes the average log training time for
    the exact ARD and RBF
    kernels across the smallest five medium datasets; Note that these methods are
    intractable on any larger datasets, hence we did not compare them
    on any of the larger datasets.
    \label{fig:sml}}
\end{figure}
%


\begin{sidewaystable}[tbhp]
  \centering
  \caption{Comparative RMSE performance on all datasets, with $n$ training points
    and $d$ the input dimensions. The results are
    averaged over 10 equal partitions of the data $\pm$ 1 standard deviation.
    We use exact RBF and ARD kernels on the small datasets, and Fastfood
    expansions of these kernels on the medium and large datasets.
    For GM and PWL, we set $Q=5$ and use
    $m=256$ sample points for each component. For FSARD and SSGPR, we set
    $m=512$. For medium datasets, we set $Q = 3$ and $m=256$ for GM and PWL.
    On the large datasets \emph{3D Road network} and \emph{Buzz}, GM and PWL
    use $Q=6$ and $m=16$, and $Q=5$ and $m=8$ for \emph{Song} and
    \emph{Household electric}.  All other methods use $Qm$ basis functions.
    The terms `small', `medium' and `large' are meant to be taken in the
    context of Gaussian process regression, which is typically intractable for
    $n > 2000$ when kernel learning is taking place.}
  \fontsize{9pt}{0.9em}\selectfont
\begin{tabular}{@{}l  r  r  r r  r  r  r  r  r}
    Datasets & n & d & RBF & ARD & FSARD & GM & PWL & FSGBARD & SSGPR \\
\hline
 \texttt{SMALL}:  \\
    Challenger  & 23 & 4 & 0.63$\pm$0.26 & 0.63$\pm$0.26 & 0.72$\pm$0.34
                         & \bf{0.61$\pm$0.27} & 0.68$\pm$0.28 & 0.77$\pm$0.39
                         & 0.75$\pm$0.38\\

    Fertility & 100 & 9 & \bf{0.19$\pm$0.04} & 0.21$\pm$0.05 & 0.21$\pm$0.05
                        & \bf{0.19$\pm$0.05} & \bf{0.19$\pm$0.05} & 0.20$\pm$0.04
                        & 0.20$\pm$0.06\\

    Slump & 103 & 7 & 4.49$\pm$2.22 & 4.72$\pm$2.42 & 3.97$\pm$2.54
                             & \bf{2.99$\pm$1.14} & 3.45$\pm$1.66
                             & 5.40$\pm$2.44 & 6.25$\pm$3.70\\

    Automobile & 159 & 15 & \bf{0.14$\pm$0.04} & 0.18$\pm$0.07 & 0.18$\pm$0.05
                          & 0.15$\pm$0.0311 & \bf{0.14$\pm$0.03}
                          & 0.22$\pm$0.08 & 0.17$\pm$0.06\\

    Servo & 167 & 4 & 0.29$\pm$0.07 & 0.28$\pm$0.09 & 0.29$\pm$0.08
                    & \bf{0.27$\pm$0.07} & 0.28$\pm$0.09 & 0.44$\pm$0.10
                    & 0.38$\pm$0.08\\

    Cancer & 194 & 34 & 32$\pm$4 & 35$\pm$4
                             & 43$\pm$9 & \bf{31$\pm$4}
                             & 35$\pm$6 & 34$\pm$5
                             & 33$\pm$5\\

    Hardware & 209 & 7 & 0.44$\pm$0.06& 0.43$\pm$0.04& 0.44$\pm$0.06
                       & 0.44$\pm$0.04& \bf{0.42$\pm$0.04}
                       & \bf{0.42$\pm$0.08} & 0.44$\pm$0.1\\

    Yacht & 308 & 7 & 0.29$\pm$0.14 & 0.16$\pm$0.11 & 0.13$\pm$0.06
                    & 0.13$\pm$0.08 & \bf{0.12$\pm$0.07} & \bf{0.12$\pm$0.06}
                    & 0.14$\pm$0.10\\

    Auto MPG & 392 & 7 & 2.91$\pm$0.30 & 2.63$\pm$0.38 & 2.75$\pm$0.43
                       & \bf{2.55$\pm$0.55} & 2.64$\pm$0.52 & 3.30$\pm$ 0.69
                       & 3.19$\pm$0.56\\

    Housing & 509 & 13 & 3.33$\pm$0.74 & 2.91$\pm$0.54 & 3.39$\pm$0.74
                       & 2.93$\pm$0.83 & \bf{2.90$\pm$0.78} & 4.62$\pm$0.85
                       & 4.49$\pm$0.69\\

    Forest fires & 517 & 12 & \bf{1.39$\pm$0.15} & \bf{1.39$\pm$0.16}
                            & 1.59$\pm$0.12 & 1.40$\pm$0.17
                            & 1.41$\pm$0.16 & 2.64$\pm$0.25
                            & 2.01$\pm$0.41\\

    Stock & 536 & 11 & 0.016$\pm$0.002 & \bf{0.005$\pm$0.001} & \bf{0.005$\pm$0.001}
                     & \bf{0.005$\pm$0.001} & \bf{0.005$\pm$0.001}
                     & 0.006$\pm$0.001 & 0.006$\pm$0.001\\

    Pendulum & 630 & 9 & 2.77$\pm$0.59 & \bf{1.06$\pm$0.35} & 1.76$\pm$0.31
                       & \bf{1.06$\pm$0.27} & 1.16$\pm$0.29 & 1.22$\pm$0.26
                       & 1.09$\pm$0.33\\

    Energy & 768 & 8 & 0.47$\pm$0.08 & 0.46$\pm$0.07
                     & 0.47$\pm$0.07 & \bf{0.31$\pm$0.07}
                     & 0.36$\pm$0.08 & 0.39$\pm$0.06
                     & 0.42$\pm$0.08\\

    Concrete & 1,030 & 8 & 5.42$\pm$0.80 & 4.95$\pm$0.77
                        & 5.43$\pm$0.76 & \bf{3.67$\pm$0.71}
                        & 3.76$\pm$0.59 & 4.86$\pm$0.97
                        & 5.03$\pm$1.35\\

    Solar flare & 1,066 & 10 & \bf{0.78$\pm$0.19} & 0.83$\pm$0.20
                            & 0.87$\pm$0.19 & 0.82$\pm$0.19
                            & 0.82$\pm$0.18 & 0.91$\pm$0.19
                            & 0.89$\pm$0.20\\

    Airfoil & 1,503 & 5 & 4.13$\pm$0.79 & 1.69$\pm$0.27 & 2.00$\pm$0.38
                       & \bf{1.38$\pm$0.21} & 1.49$\pm$0.18 & 1.93$\pm$0.38
                       & 1.65$\pm$0.20\\

    Wine & 1,599 & 11 & 0.55$\pm$0.03 & \bf{0.47$\pm$0.08} & 0.50$\pm$0.05
                     & 0.53$\pm$0.11 & 0.48$\pm$0.03 & 0.57$\pm$0.04
                     & 0.66$\pm$0.06\\
    \hline
     \texttt{MEDIUM}:  \\
    Gas sensor & 2,565 & 128 & 0.21$\pm$0.07 & \bf{0.12$\pm$0.08} & 0.13$\pm$0.06
                            & 0.14$\pm$0.08 & \bf{0.12$\pm$0.07}
                            & 0.14$\pm$0.07 & 0.14$\pm$0.08\\

    Skillcraft & 3,338 & 19 & 1.26$\pm$3.14 & \bf{0.25$\pm$0.02}
                           & \bf{0.25$\pm$0.02} & \bf{0.25$\pm$0.02}
                           & \bf{0.25$\pm$0.02} & 0.29$\pm$0.02
                           & 0.28$\pm$0.01\\

    SML & 4,137 & 26 & 6.94$\pm$0.51 & 0.33$\pm$0.11 & \bf{0.26$\pm$0.04}
                    & 0.27$\pm$0.03 & 0.31$\pm$0.06 & 0.31$\pm$0.06
                    & 0.34$\pm$0.05\\

    Parkinsons & 5,875 & 20 & 3.94$\pm$1.31 & 0.01$\pm$0.00 & 0.02$\pm$0.01
                           & \bf{0.00$\pm$0.00} & 0.04$\pm$0.03
                           & 0.02$\pm$0.00 & 0.08$\pm$0.19\\

    Pumadyn & 8,192 & 32 & 1.00$\pm$0.00 & 0.20$\pm$0.00 & 0.22$\pm$0.03
                        & 0.21$\pm$0.00 & \bf{0.20$\pm$0.00} & 0.21$\pm$0.00
                        & 0.21$\pm$0.00\\

    Pole Tele & 15,000 & 26 & 12.6$\pm$0.3 & 7.0$\pm$0.3
                              & 6.1$\pm$0.3 & 5.4$\pm$0.7
                              & 6.6$\pm$0.3 & 4.7$\pm$0.2
                              & \bf{4.3$\pm$0.2}\\

    Elevators  &    16,599 & 18 & 0.12$\pm$0.00 & 0.090$\pm$0.001
                               & 0.089$\pm$0.002 & 0.089$\pm$0.002
                               & 0.089$\pm$0.002 & \bf{0.086$\pm$0.002}
                               & 0.088$\pm$0.002\\

    Kin40k & 40,000 & 8 & 0.34$\pm$0.01 & 0.28$\pm$0.01 & 0.23$\pm$0.01
                            & 0.19$\pm$0.02 & 0.23$\pm$0.00 & 0.08$\pm$0.00
                            & \bf{0.06$\pm$0.00}\\

    Protein & 45,730 & 9 & 1.64$\pm$1.66 & 0.53$\pm$0.01 & 0.52$\pm$0.01
                        & 0.50$\pm$0.02 & 0.52$\pm$0.01 & 0.48$\pm$0.01
                        & \bf{0.47$\pm$0.01}\\

    KEGG & 48,827 & 22 & 0.33$\pm$0.17 & \bf{0.12$\pm$0.01}
                      & \bf{0.12$\pm$0.01} & \bf{0.12$\pm$0.01}
                      & \bf{0.12$\pm$0.01} & \bf{0.12$\pm$0.01}
                      & \bf{0.12$\pm$0.01}\\

    CT slice & 53,500 & 385 & 7.13$\pm$0.11 & 4.00$\pm$0.12 & 3.60$\pm$0.09
                           & 2.21$\pm$0.06 & 3.35$\pm$0.08 & 2.56$\pm$0.12
                           & \bf{0.59$\pm$0.07}\\

    KEGGU & 63,608 & 27 & 0.29$\pm$0.12 & \bf{0.12$\pm$0.00}
                       & \bf{0.12$\pm$0.00} & \bf{0.12$\pm$0.00}
                       & \bf{0.12$\pm$0.00} & \bf{0.12$\pm$0.00}
                       & \bf{0.12$\pm$0.00}\\
    \hline
     \texttt{LARGE}:  \\
    3D road & 434,874 & 3 & 12.86$\pm$0.09 & 10.91$\pm$0.05
                         & 10.29$\pm$0.12 & 10.34$\pm$0.19
                         & 11.26$\pm$0.22 & \bf{9.90$\pm$0.10}
                         & 10.12$\pm$0.28 \\

    Song & 515,345 & 90 & 0.55$\pm$0.00 & 0.49$\pm$0.00 & 0.47$\pm$0.00
                       & 0.46$\pm$0.00 & 0.47$\pm$0.00 & 0.46$\pm$0.00
                       & \bf{0.45$\pm$0.00}\\

    Buzz & 583,250 & 77 & 0.88$\pm$0.01 & 0.59$\pm$0.02 & 0.54$\pm$0.01
                       & \bf{0.51$\pm$0.01} & 0.54$\pm$0.01 & 0.52$\pm$0.01
                       & 0.54$\pm$0.01\\

    Electric & 2,049,280 & 11 & 0.23$\pm$0.00 & 0.12$\pm$0.12 & 0.06$\pm$0.01
                            & \bf{0.05$\pm$0.00} & 0.07$\pm$0.04
                            & \bf{0.05$\pm$0.00} & \bf{0.05$\pm$0.00}\\
\end{tabular}
\label{table: testresults}
\end{sidewaystable}

\begin{figure}[htb]
  \centering
  \includegraphics[width=0.49\textwidth]{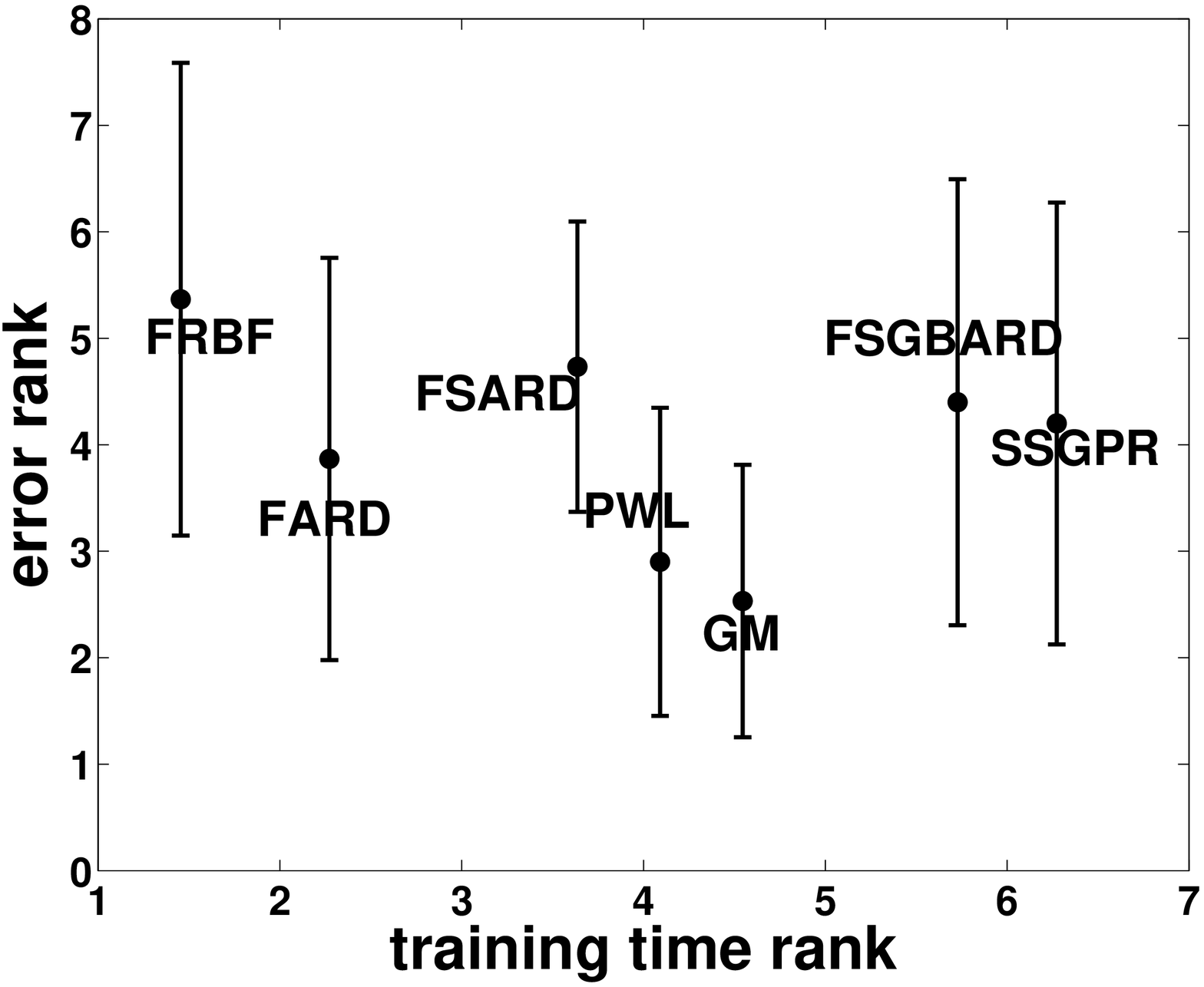}
  \includegraphics[width=0.49\textwidth]{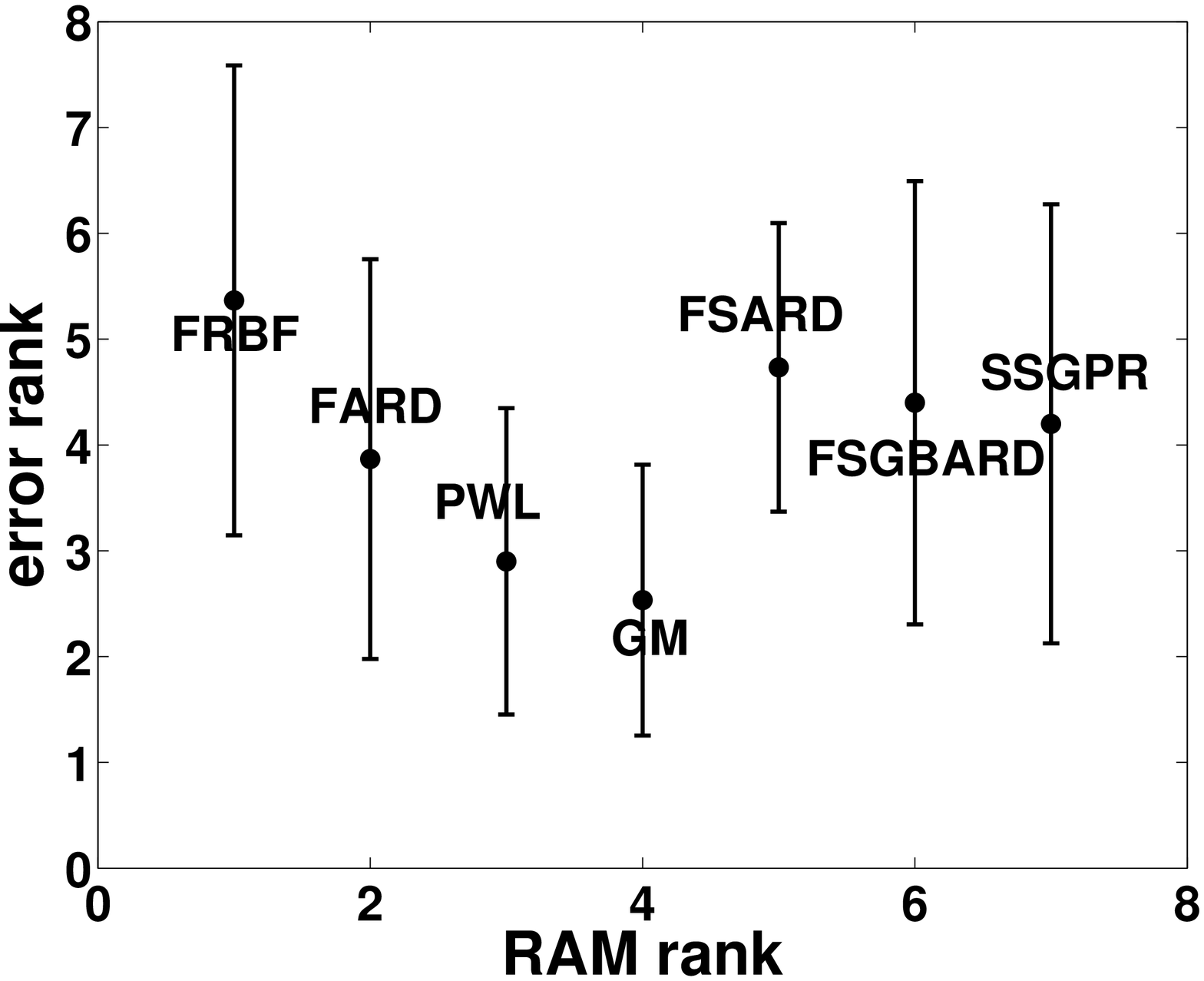}
  \caption{\label{fig:rank}%
    To assess the relative performance of the algorithms with respect
    to another, we compare their relative error rank as a function of
    training time (left) and memory footprint (right).} %
\end{figure}

\begin{figure}[htb]
  \centering
  \includegraphics[width=0.49\textwidth]{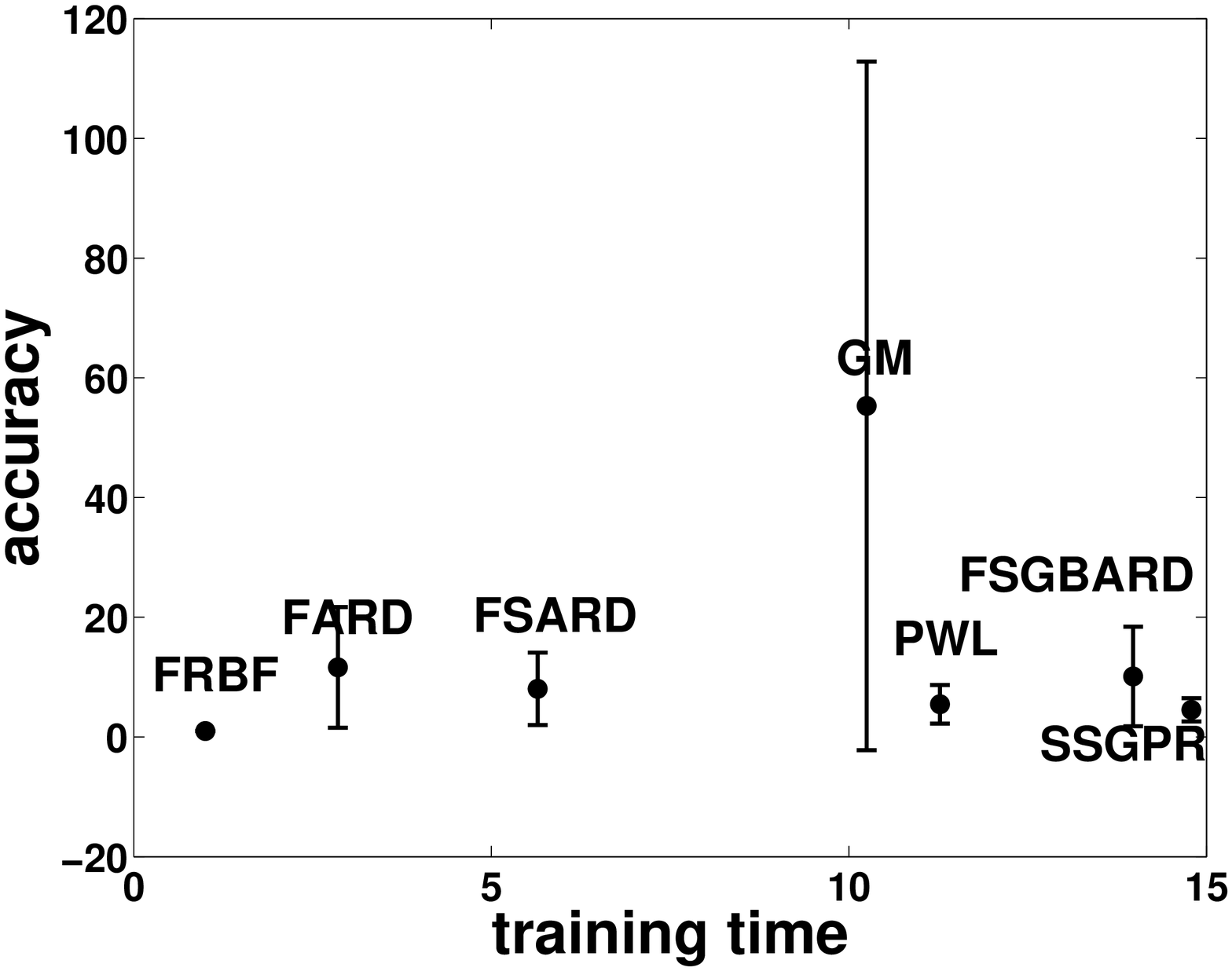}
  \includegraphics[width=0.49\textwidth]{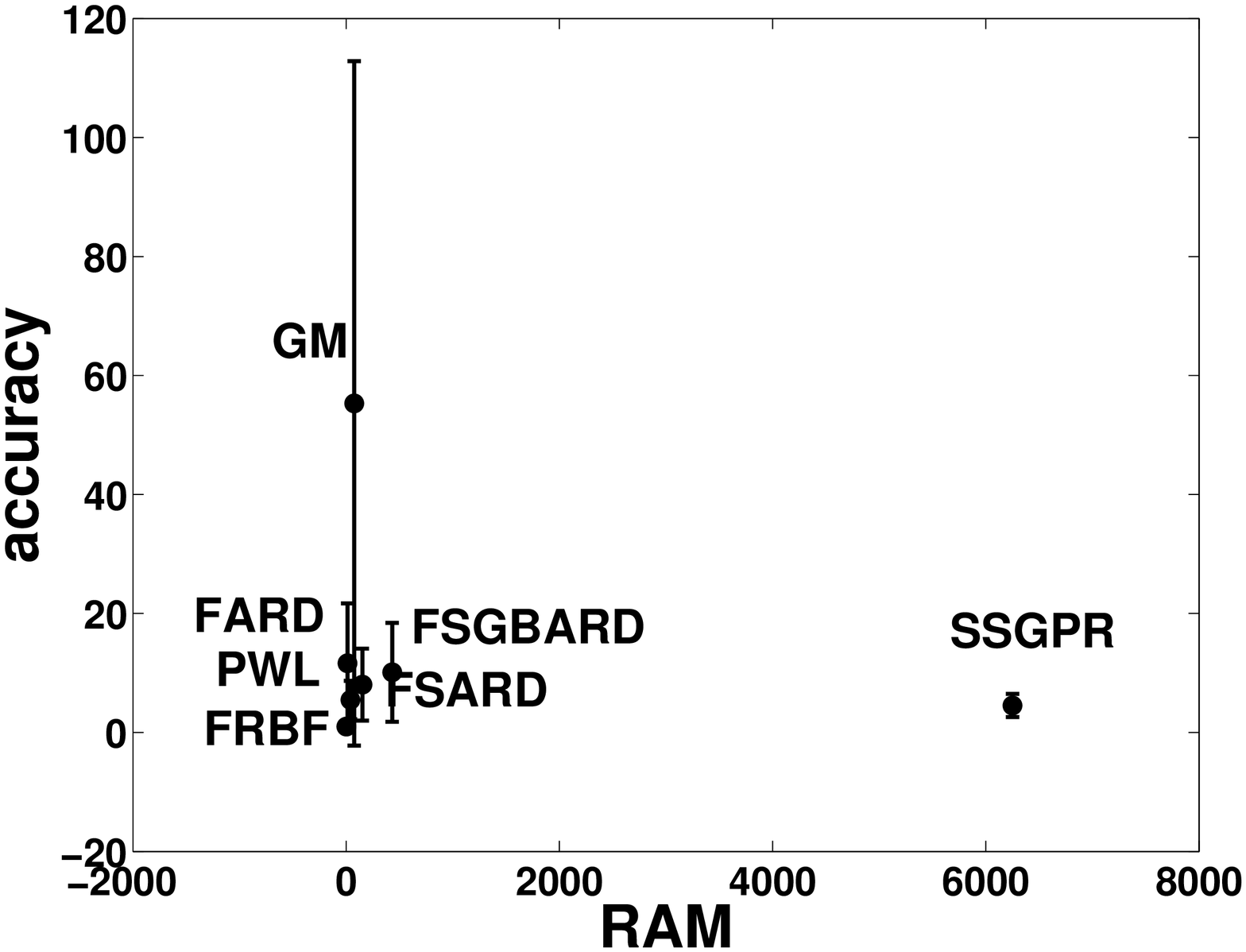}
  \caption{\label{fig:cmpnolog}%
    Accuracy of the algorithms as a function of training time
    (left) and memory (right).}
\end{figure}

We use the {\em same} number of basis functions for all methods. We use $Q$ to
denote the number of components in GM and PWL and $m$ to denote the number of
basis functions in each component. For all other methods, we use $Qm$
basis functions.  For the largest datasets in Table \ref{table: testresults} we
favoured larger values of $Q$, as the flexibility of having more components $Q$
in GM and PWL becomes more valuable when there are many datapoints; although
we attempted to choose sensible $Q$ and $m$ combinations for a particular model
and number of datapoints $n$, these parameters were not fine tuned.  We choose
$Qm$ to be as large as is practical given computational constraints, and SSGPR
is allowed a significantly larger parametrization.

Indeed SSPGR is allowed $Qmd+2$ free parameters to learn, and we set $Q \ll m$.
This setup gives SSGPR a significant advantage over our proposed
models.  However, we
wish to emphasize that the GM, PWL, and FSGBARD models are competitive with
SSGPR, even in the adversarial situation when SSGPR has many orders of magnitude more free parameters than GM or PWL.  For comparison, the RBF, ARD, PWL, GM, FSARD and FSGBARD methods respectively require $3, d+2, Q(d+3)+1, Q(2d+1)+1, Qm+d+2,$ and $3Qm+d+2$ hyperparameters.

Gaussian processes are most commonly implemented with exact RBF and ARD kernels, which we run on the smaller ($n < 2000$) datasets in Table \ref{table: testresults},
where the proposed GM and PWL approaches generally perform better than all alternatives.  On the larger datasets, exact ARD and RBF kernels are entirely intractable, so we compare to Fastfood expansions.  That is, GM and PWL are both more expressive and profoundly more scalable than exact ARD and RBF kernels, far and above the most popular alternative approaches.

In Figure \ref{fig:qbasis} we investigate how RMSE
performance changes as we vary $Q$ and $m$. The GM and PWL models continue to increase in
performance as more basis functions are used.  This trend is not present with SSGPR or
FSGBARD, which unlike GM and PWL, becomes more susceptible to over-fitting as
we increase the number of basis functions.  Indeed, in SSGPR, and in FSGBARD
and FSARD to a lesser extent, more basis functions means more parameters to
optimize, which is not true with the GM and PWL models.

To further investigate the performance of all methods, we compare each of the
seven tested methods over all experiments, in terms of average normalised log
predictive accuracy, training time, testing time, and memory consumption, shown
in Figures \ref{fig:accutimelog} and \ref{fig:testtime} (higher accuracy scores and lower training time, test
time, and memory scores, correspond to better performance). Despite the reduced
parametrization, GM and PWL outperform all alternatives in accuracy, yet
require similar memory and runtime to the much less expressive FARD model, a
Fastfood expansion of the ARD kernel.  Although SSGPR performs
third best in accuracy, it requires more memory, training time, testing
runtime (as shown in Fig~\ref{fig:testtime}), than all other models.  FSGBARD
performs similar in accuracy to SSGPR, but is significantly more time and
memory efficient, because it leverages a Fastfood representation.  For clarity,
we have so far considered log plots.  If we view the results without a log
transformation, as in Fig~\ref{fig:cmpnolog} (supplement) we see that GM and SSGPR are
outliers: on average GM greatly outperforms all other methods in predictive
accuracy, and SSGPR requires profoundly more memory than all other methods.

\section{Discussion}
\label{sec:blah}

Kernel learning methods are typically intractable on large datasets, even
though their flexibility is most valuable on large scale problems.  We have
introduced a family of flexible, scalable, general purpose and lightly
parametrized kernel methods, which learn the properties of groups of spectral
frequencies in Fastfood basis function expansions.  We find, with a minimal
parametrization, that the proposed methods have impressive performance on a
large and diverse collection of problems -- in terms of predictive accuracy,
training and test runtime, and memory consumption.  In the future, we expect
additional performance and efficiency gains by automatically learning the
relative numbers of spectral frequencies to assign to each group.

In short, we have shown that we can have kernel methods which are
simultaneously scalable and expressive.  Indeed, we hope that this work will
help unify efforts in enhancing scalability and flexibility for kernel methods.
In a sense, flexibility and scalability are one and the same problem: we want
the most expressive methods for the biggest datasets.

\bibliographystyle{plainnat}
\bibliography{../../../bibfile}

\end{document}